\documentclass[10pt,twocolumn,letterpaper]{article}

\usepackage{cvpr}

\usepackage{graphicx,subcaption}
\usepackage{comment}
\usepackage{amsmath}
\usepackage{amssymb}
\usepackage{booktabs}
\usepackage[super]{nth}

\usepackage[pagebackref,breaklinks,colorlinks]{hyperref}
\hypersetup{colorlinks,urlcolor=blue}

\usepackage[capitalize]{cleveref}
\crefname{section}{Sec.}{Secs.}
\Crefname{section}{Section}{Sections}
\Crefname{table}{Table}{Tables}
\crefname{table}{Tab.}{Tabs.}

\def\cvprPaperID{*****} 
\def\confName{CVPR}
\def\confYear{2022}

\begin{document}

\title{FedOS: using open-set learning to stabilize training in federated learning}

\author{Mohamad Mohamad\\
{\tt\small \nolinkurl{mohamad_m.2@live.com}}
\and
Julian Neubert\\
{\tt\small \nolinkurl{julian.neubert@gmx.de}}
\and
Juan Segundo Argayo\\
{\tt\small \nolinkurl{jargayo@gmail.com}}
}
\maketitle

\begin{abstract}
   Federated Learning is a recent approach to train statistical models on distributed datasets without violating privacy constraints. The data locality principle is preserved by sharing the model instead of the data between clients and the server. This brings many advantages but also poses new challenges. In this report, we explore this new research area and perform several experiments to deepen our understanding of what these challenges are and how different problem settings affect the performance of the final model. Finally, we present a novel approach to one of these challenges and compare it to other methods found in literature. The code is available on our github \footnote{\url{https://github.com/mohamad-m2/federated-learning}}.
\end{abstract}

\section{Introduction}

Nowadays, phones are the primary and sometimes the only computing device for many people. As such, the interaction with them and their powerful sensors leave an unprecedented amount of rich data, much of it private in nature. Cameras, microphones, GPS, and tactile sensors are all examples of this. In the field of Big Data and Machine Learning applications, this data regains importance as models learned with it hold the promise of greatly improving usability by powering more intelligent applications, but the sensitive nature of the data means there are risks and responsibilities to storing it in a centralized location \cite{mcmahan2017communication}.

Federated learning (FL) is a privacy-preserving framework, originally introduced by McMahan \etal \cite{DBLP:journals/corr/McMahanMRA16}, for training models from decentralized user data residing on devices at the edge, the one mentioned before. Models are trained iteratively across many federated rounds. For each round, every participating device (a.k.a. client), receives an initial model from a central server, performs stochastic gradient descent (SGD) on its local training data, and sends back the gradients. The server then aggregates all gradients from the participating clients and updates the starting model \cite{hsu2020fedir}. In FL, raw client data is never shared with the server or other clients. This distinguishes FL from traditional distributed optimization and requires dealing with heterogeneous data

\section{Related Work}

\begin{table*}
  \centering
  \begin{tabular}{ | l | c | c | c | c | c | c | c | c | c | c | }
    \hline
    lr & 0.2 & 0.15 & 0.1 & 0.08 & 0.06 & 0.05 & 0.04 & 0.02 & 0.01 & 0.005 \\
    \hline
    raw & 47.67 & 44.7 & 49.26 & 47.75 & 49.98 & 48.61 & 47.46 & 44.13 & 43.83 & 39.92 \\
    scaled & 51.08 & 49.88 & 53.33 & 52.29 & 48.09 & 50.77 & 51.27 & 43.80 & 37.21 & 28.71 \\
    standardized & 53.42 & 54.07 & 54.27 & 53.53 & 53.85 & 52.32 & 52.74 & 47.54 & 40.56 & 34.81 \\
    \hline
  \end{tabular}
  \caption{Hyper-parameter tuning on centralized setting. Learning rate and pre-processing.}
  \label{tab:hyperparams}
\end{table*}

\textbf{FedProx.} There are two key challenges in Federated Learning that differentiate it from traditional distributed optimization: (1) significant variability in terms of the system's characteristics on each device in the network (systems heterogeneity), and (2) non-identically distributed data across the network (statistical heterogeneity). FedProx (Tian Li \etal \cite{https://doi.org/10.48550/arxiv.1812.06127}) can be viewed as a generalization and re-parametrization of FedAvg. In this work, the authors have theoretically proven convergence guarantees when learning over data from non-identical distributions (satistical heterogeneity), while also adhering to the constraints each device-level system has by allowing each participating device to perform a variable amount of work (system heterogeneity). They demonstrate that in highly heterogeneous settings, FedProx has significantly more stable and accurate convergence behavior relative to FedAvg, improving absolute test accuracy by 22\% on average.

\textbf{FedIR.} As well as \cite{https://doi.org/10.48550/arxiv.1812.06127}, the authors of this paper (Hsu \etal \cite{hsu2019fedavgm}) recognized that one of the challenges in terms of data diversity relies on the fact that data at the source is far from independent and identically distributed (IID). In addition, they also found that differing quantities of data are typically available at each device (imbalance). In their work, they characterized the effect these real-world data distributions have on distributed learning, using FedAvg as a benchmark. They proposed two new algorithms, FedVC and FedIR. The latter one addresses the issue of the non-identical class distribution shift present in the federated clients. If we consider a target distribution $p(x,y)$ of images $x$ and class labels $y$ on which a model is supposed to perform well, and a predefined loss function $l(x,y)$, the objective of learning is to minimize the expected loss $E_{p}[l(x,y)]$ with respect to the target distribution $p$. Instead, training examples on a federated client $k$ are sampled from a client-specific distribution $q_{k}(x,y)$. This implies that the empirical loss being optimized on every client is a \textit{biased} estimator of the loss with respect to the target distribution, since $E_{q_{k}}[l(x,y)] \neq E_{p}[l(x,y)]$. Therefore, this algorithm presents an importance reweighting scheme that applies importance weights $w_{k}(x,y)$ to every client's local objective. With those in place, an unbiased estimator of loss with respect to the target distribution can be obtained using training examples from the client distribution. The paper shows a consistent improvement versus the experiments run over the FedAvg baseline.

\section{Baseline}

In this section, we will describe the baseline we compare our experiments to and give a brief justification for some of the initial hyperparameter choices.

\subsection{Centralized Setting}
Following the project proposal, we choose LeNet-5 \cite{726791} as the architecture for our experiments, the only change is that our model takes the 3 input channels from RGB color images. This model is not the state of the art for the CIFAR-10 dataset \cite{Krizhevsky09learningmultiple}, but it is sufficient to show the relative performance of our experiments. Additionally, using such a compact model in the federated setting brings several benefits due to the potentially limited performance of client devices and the need to share the model and its updates between the server and clients for each round.

 \begin{figure*}
 \centering
 \includegraphics[width=0.8\textwidth]{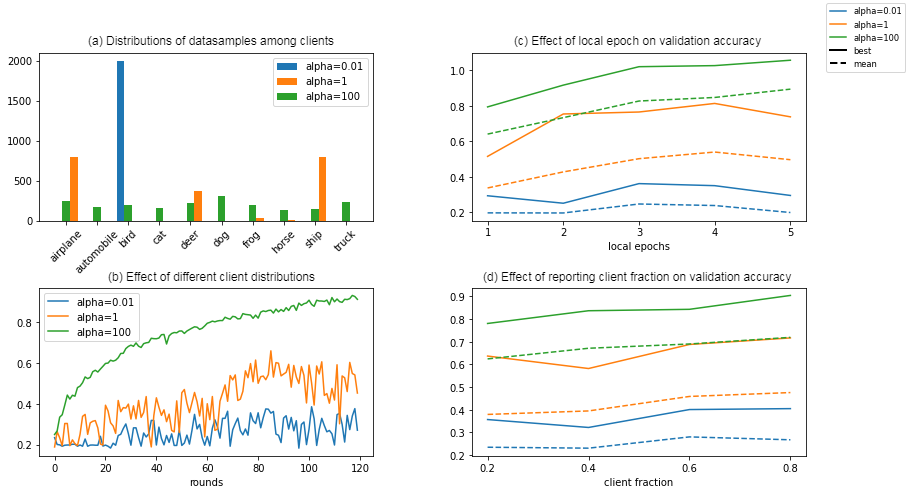}
   
    \caption{\textbf{FedAvg ablation study}. Figure (a) shows how classes are distributed for different values of $\alpha$ taking a random client as an example. Figure (c) illustrates the relative validation accuracy of FedAvg during the training process for different values of $\alpha$. Figures (b) and (d) show the relative validation accuracy after 120 rounds of training with different hyperparameter settings hyperparameters by varying local epochs and reporting client fraction respectively.}
    \label{fig:fedavg_ablation}
\end{figure*}

We perform a small hyperparameter search over the learning rate and pre-processing to find an optimal setting to conduct our later experiments. The results of which can be seen in \Cref{tab:hyperparams}. We notice, that standardizing the data yields the best results. Interestingly, however, the model performed equally well over a wide range of learning rates. The decreased performance of models trained with lower learning rates is probably due to the short training process. They would likely converge after more epochs to similarly high values of accuracy, if not more. A learning rate schedule could help us circumvent this problem. However, for simplicity, we choose a learning rate of $0.1$ for all following experiments. In this setting, we achieve a final validation accuracy of $54.03\%$ after 120 epochs of training.

\subsection{Federated Setting}

Similar to Hsu \etal \cite{hsu2019fedavgm} we draw the data distributions for our clients from a Dirichlet distribution. This parametric distribution allows us to compare how different federated approaches handle different severities of unbalanced client data by varying $\alpha$. For very large $\alpha$, the clients will have an almost equal distribution among the classes. The smaller the value for $\alpha$, the more extreme the difference between the clients. Having $\alpha=0$ will lead to each client only having samples from a single class. For now, we will set $\alpha=1$ and further explore the effect of different client distributions in our ablation studies in \cref{sec:ablation}. Having clients with diverse data is important for our experiments, as this difference in data availability on each client device is one of the major limiting factors for federated learning when compared to a centralized baseline. Additionally, it mirrors how data is available in the real world, with different users having different interests, therefore gathering data about different topics, in different quantities and qualities. As we can not emulate all of these parameters with the given dataset, we focus our experiments on the distribution of classes among clients.

Furthermore, we keep our baseline simple by implementing a synchronous communication scheme and using federated averaging to aggregate the clients' updates. \Cref{sec:experiments} shows some experiments with more advanced aggregation schemes. By default, we distribute the dataset among $20$ clients, all with equal number of samples ($2000$). For each communication round, $20\%$ of them are chosen to train the network for one epoch on their local data and report their new weights back to the server. We evaluate our results by comparing the final model's accuracy on a centralized validation set held by the server. This allows us to analyze how well the model was able to train on the distributed data and provides a more accurate comparison to its centralized counterpart. This is functionally equivalent to comparing the weighted accuracy between clients if the total test set distributed among the clients is equal to the central test set and all clients report their local accuracy.

In this setting, we achieve a relative validation accuracy of $45.98\%$ after 120 communication rounds and $\alpha=1$.

\section{Ablation study}
\label{sec:ablation}

Since we choose the hyperparameters for our baseline without any factual basis, we will use this section to explore the effect each parameter has on the final result of the model.

\subsection{Data distribution}

Probably the most important factor affecting validation accuracy is the data distribution on each client device. Most machine learning methods assume the data they train on to be independent and identically distributed (IID). This means all data samples are drawn from the same distribution and independently from each other. Violating this assumption often leads to decreased performance of the model.

\Cref{fig:fedavg_ablation} (a) shows the number of samples a typical client has access to given different values of $\alpha$. We observe that the smaller $\alpha$ is, the stronger the class unbalance on each client. For very small values such as $\alpha=0.01$, each client typically only has access to samples of a single class. This violation of the IID assumption leads to strong performance decreases. \Cref{fig:fedavg_ablation} (c) shows the relative validation accuracy during training over the number of communication rounds. For $\alpha=100$ the data is more or less evenly distributed among the clients, so the model can achieve results very similar to those of the centralized case, albeit after a longer training process. For $\alpha=1$ and lower values, the model does not only produce much worse results, but we also observe that the training is highly unstable for these hard cases.

 \begin{figure*}[t]
 \centering
 \includegraphics[width=1\textwidth]{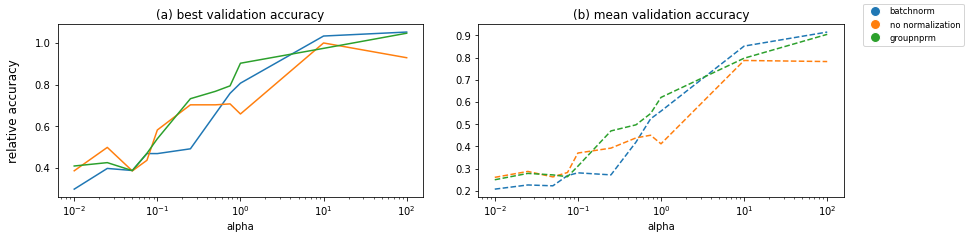}
   
    \caption{\textbf{Normalization effect}. we show in these graphs how different normalization layers perform compared to the raw model for various $\alpha$ values. We present both mean and best validation accuracy.}
    \label{fig:normalization}
\end{figure*}

\subsection{Client settings}

Since the homogeneity of the data distribution on each client device is typically dictated by the task, it is worth analyzing which other parameters exist, that affect the predictive quality of the final model, and that can be influenced by the application. Here we analyze the effects of how many epochs the local modal is trained for on each selected client device during each round of communication and the fraction of clients chosen to participate in each round of communication relative to the total number of clients. In response to the inconsistent validation accuracy during the training process, that we discussed in the previous subsection, we will report both the average and the best validation accuracy each model achieved over the training process.

\Cref{fig:fedavg_ablation} (b) shows what happens when we increase the number of local epochs. We observe that this helps for large values of $\alpha$, while it seems to have no effect for smaller values. An intuitive explanation for this could be, that increasing the number of local epochs results in a longer effective training process, since the number of communication rounds does not change. However, this effect only helps in the case of more homogeneous data distributions on the clients, since each model is expected to have similar gradients. If the clients' data grows more heterogeneous, each model will converge towards a different local optimum, rendering the weight updates, that are sent back to the server incompatible.

Another parameter of the training process we experiment with is the relative amount of clients, that are chosen to train on their local data for each round. \Cref{fig:fedavg_ablation} (d) shows the results of this study. While the results do not differ significantly from each other, we do see a slight improvement for larger client fractions. This remains true over all values of $\alpha$, but more pronounced for larger values. Choosing a larger fraction of clients to train each round most likely stabilizes training, as the average update step will more likely point towards the true local minimum of the loss function. Again, this technique is more effective for larger values of $\alpha$, as the weight updates of the clients are more compatible with each other.

\section{Normalization}
Since normalization techniques proved their capabilities in helping deep networks train faster and attain better results, they have been widely used in many benchmark models \cite{DBLP:journals/corr/HeZRS15,https://doi.org/10.48550/arxiv.1706.03762}. While batch normalization \cite{DBLP:journals/corr/IoffeS15} helps to minimize internal covariate shift by batch statistics, it runs into problems with small batches. This issue is addressed with an alternative normalization method known as Group normalization \cite{DBLP:journals/corr/abs-1803-08494}. Group Normalization uses channel instead of batch statistics, which makes it more robust to the challenges and limitations that come with small batch sizes. Considering how widely both techniques used are, we discuss here how they perform in the federated scenario. 

\subsection{Batch normalization}
As mentioned earlier, batch normalization uses batch statistics to compute a mean and a variance used for standardizing the batch, and then applying an affine transformation with learnable parameters. Usually, the shortcoming of this technique is its inaccurate estimation when having batches of insufficient sizes. In our study, we notice that Batch Normalization boosts the model's learning capabilities while $\alpha$ is large but these improvements aren't achievable anymore when $\alpha$ decreases below a specific threshold. In fact for low $\alpha$ values Batch Normalization hurts the training of the model (Figure \ref{fig:normalization}) and attains worse results than models with no normalization layers. This is mainly due to its biased estimation but in this case, the problem isn't the batch size, instead, it is the highly heterogeneous batches between clients. This results in highly biased statistics to be aggregated on the server's side.

\subsection{Group normalization}
Group normalization was introduced as a simple alternative to Batch Normalization. Group Normalization normalizes layer outputs based on channel statistics. Specifically for a number of groups $G$ ($G$ is a hyperparameter), it standardizes each group of channels' features. Group Normalization shares a similar concept as other normalization methods (layer normalization \cite{https://doi.org/10.48550/arxiv.1607.06450}, and instance normalization \cite{DBLP:journals/corr/UlyanovVL16}). Actually, in the extreme cases of $G=1$, and $G=N$ (where $N$ is the number of channels) it is equivalent to Layer Normalization and Instance Normalization respectively. We performed our analysis with (2 groups for the \nth{1} Convolutional layer with 6 channels, 4 groups for the \nth{2} one with 16 channels, and 30 groups for the last one with 120 channels), as shown in \ref{fig:normalization}, Group Normalization outperforms Batch Normalization for lower $\alpha$. This shows that group norm isn't as limited as Batch Normalization with the batch samples' distribution, still the model suffer from the non-identicalness of the client's data issue.

\begin{figure*}[t]
  \centering
  
  \begin{subfigure}[t]{0.5\linewidth}
   \includegraphics[width=\textwidth]{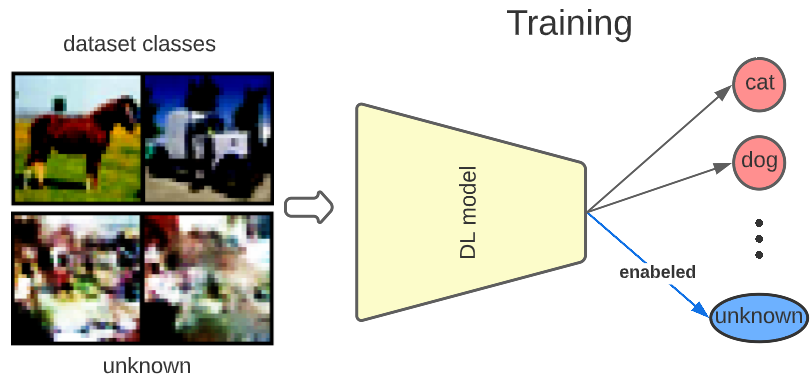}
   
    \caption{}
    \label{fig:short-a}
  \end{subfigure}
  \hfill
  \begin{subfigure}[t]{0.45\linewidth}
     \includegraphics[width=\textwidth]{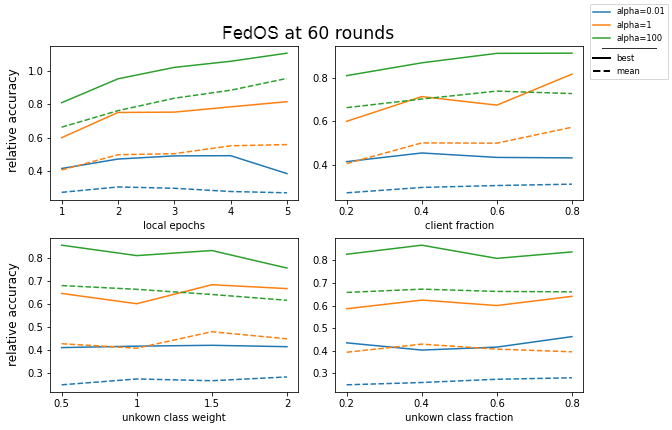}
    \caption{}
    \label{fig:short-b}
  \end{subfigure}
  \caption{\textbf{FedOS}. Figure (a) shows the structure of the model during training and some samples of the unknown class generated by the GAN. In (b) we show the results of the ablation study performed on our proposed FedOS method, in particular we show the max and mean relative accuracy over 60 rounds. The mean represents model learning stability while the max shows goodness. We have 4 hyperparameters in total with the default values (local epochs: 1, client fraction: 20\%, unknown class loss weight: 1, unknown class fraction: 60\%). For each study we change one parameter while fixing the others to their default values. }
  \label{fig:short}
\end{figure*}

\section{Our method}

As we saw in our ablation studies in \cref{sec:ablation}, one of the biggest challenges in federated learning is the heterogeneity of the client devices. During all of our experiments, the model struggled to achieve meaningful results in settings with highly unbalanced data distributions on client devices. We hypothesize that an important factor is, that most clients are only aware of a fraction of the classes the model is designed to recognize. In fact, in the case of $\alpha=0.01$, many clients only have access to a single class and are not aware of features that could be necessary to recognize other classes. This does not only pose an issue for local training on the client devices. Since each client trains only on a small subset of classes, their weights will likely diverge from each other, making the aggregation of client updates even more challenging. In this section, we present a novel approach that is designed to address these shortcomings. During the local training step on each client device, we add artificial samples from an unknown class \cite{DBLP:journals/corr/BendaleB15} that represent those classes that the client might not be aware of. Ideally, these artificial samples share the same feature space as the real images in the dataset. This allows the local model on each client to improve its feature extraction capabilities while simultaneously learning which features are important to discriminate those classes, which the local model has access to.

Let $\mathcal{D}$ denote our full dataset, $\mathcal{D_{C}}$ is the fraction of $\mathcal{D}$ belonging to Client $\mathcal{C}$, $\mathcal{\overline{D_{C}}}$ represents $\mathcal{D\setminus D_{C}}$, and $\mathcal{M_{C}}$ is the model currently training on $\mathcal{D_{C}}$. Suppose a global dataset containing all classes as $\mathcal{D_{G}}$. by definition, a class belonging to $\mathcal{D}$ belongs to $\mathcal{D_{G}}$ which implies $\mathcal{\overline{D_{C}} \subset D \subset D_{G} }$. What we want to achieve is $\mathcal{M_{C}}$ predicting unknown for $\mathcal{\overline{D_{C}}}$ for all clients so that at the aggregation step all models will distinguish their classes from those of other clients. We estimate $\mathcal{D_{G}}$ using a relatively large dataset containing a fair amount of different classes and a GAN \cite{gan} learning its general distribution. We also modify our model to accept one extra class "unknown" which introduces a new term to the loss function that symbolizes the loss of unknown samples' misclassification, this term is weighted using a hyperparameter $\mathcal{W_{U}}$.
 \begin{equation}\label{eqn:loss}
  L(C)= L(D_{C})+W_{U}* L(\text{unknown})
  \end{equation}

\subsection{Pre-training}
Inspired by \cite{Neal_2018_ECCV,ge2017generative} we train our GAN for generating and modeling the unknown class. We differ from them in the final objective we have for training the GAN. In the context of open-set recognition, the model at the end should be able to recognize high-resolution unknown samples for which it has never been trained, so they train their adversarial network to model the out-of-distribution space using the classes at hand. On the other hand, our objective is to take into account the existence of numerous different classes that don't appear within the local data during training. We try to model the general space features by training the network on many classes to capture the most repeated features. We train our GAN on a random subset of the Tiny Imagenet dataset \cite{Le2015TinyIV} containing samples from different classes. this step has a low cost since we don't need a specific class to train on, which makes the training almost unsupervised. The only requirement is to train on many classes. 

\subsection{Training}
First, the trained GAN is sent to the client-side, this step accounts only for half a round since the adversarial network doesn't need to be sent back to the server-side and its size is comparable to the model used for classification. The clients generate a set of random samples proportional to the amount of data they have locally. This is controlled by our second hyper parameter $\mathcal{F_{U}}$. Suppose  $\mathcal{N}$ is the total number of local samples, and the quantity of unknown samples generated is $\mathcal{F_{U}* N}$.
We add an extra output class to classify "unknown" samples (see Figure \ref{fig:short-a}) and the training proceeds with the normal FedAvg scheme. At the end of training and before inference, we disable the extra output forcing the model to output one of the known classes since our final objective is a closed set classification task.

\begin{table*}[t]
\centering
\begin{tabular}{c c c c c c c c c }

\multicolumn{1}{c }{\textbf{}} &
\multicolumn{8}{c }{\textbf{Relative accuracy @N rounds}  }\\
\toprule

\multicolumn{1}{c }{\textbf{}} & \multicolumn{4}{c }{\textbf{Alpha=1}} & \multicolumn{4}{c }{\textbf{Alpha=0.01}} \\
\cmidrule(l){2-5}
\cmidrule(ll){6-9}
\multicolumn{1}{ c }{\textbf{Methods}} & \textbf{@125} & \textbf{@250} & \textbf{@375} & \textbf{@500} & \textbf{@125} & \textbf{@250} & \textbf{@375} & \textbf{@500} \\ 
\cmidrule(l){1-1}
\cmidrule(l){2-5}
\cmidrule(ll){6-9}
\multicolumn{1}{ c }{FedAvg}  & 65.94 & 82.17 & 83.03 & 88.34 & 40.41 & 41.92 & 51.19 & 60.75 \\ 
\multicolumn{1}{ c }{FedIR}& 77.66 & 82.98 & 89.03 & 89.36 &  \textbf{51.75} & \textbf{59.56} & 64 & 69.03 \\ 
\multicolumn{1}{ c }{FedProx} & 63.85 & 82.07 & 86.51 & 91.4  & 28.81 & 37.64 & 48.58 & 48.58 \\ 
\multicolumn{1}{ c }{FedOS} & \textbf{77.92} & \textbf{84.09} & \textbf{94.98} & \textbf{95.85} & 46.6 & 56.13 & \textbf{64.7} & \textbf{69.11} \\ 

\bottomrule
\end{tabular}
\caption{Performance comparison of different methods}
\label{tab:results}
\end{table*}

\subsection{FedOS ablation study}
To study how the different hyperparameters affect our model behavior we conduct a detailed ablation study on all 4 parameters (local epochs $E$, reporting client fraction $CL$, unknown class
weight $\mathcal{W_{U}}$, and unknown class fraction $\mathcal{F_{U}}$). For each of them, we perform tests varying only one value while keeping all other parameters constant with default settings (refer to \ref{fig:short} for the default values). For high homogeneity levels ($\alpha=100$) the method behaves similarly to FedAvg \cref{fig:fedavg_ablation} where increasing local epochs or client fraction boosts performance. On the other side varying the parameters controlling the unknown class doesn't have the same effect simply since at high values of $\alpha$ the new unknown class isn't strictly needed. It is still interesting to point out that for $\mathcal{W_{U}}=0.5$ the accuracy is higher than FedAvg with the same settings, which shows that the additional class could have a regularizing effect. Moving down to $\alpha=1$ we still have the same effect for $E$ and $CL$, however, for $\mathcal{W_{U}}=1.5$ we have better performance and stability (max and mean accuracy). Similarly for increasing $\mathcal{F_{U}}$ we obtain better results. Setting $\mathcal{F_{U}}=0.8$ is overkill. Even if it has a slightly better performance than 0.4, it has decreased training stability. The most interesting result is at $\alpha=0.01$ where our method is at its full potential. The method outperforms FedAvg results for hyperparameter settings. Performing a few additional local epochs further enhances performance, it does, however, drop drastically at $E=5$. Similar behavior occurs with $CL$, but contrary to $E$, model stability keeps improving. Changing $\mathcal{W_{U}}$ seems to have minor effects for this $\alpha$, however, choosing the highest $\mathcal{F_{U}}$ boosts both stability and performance in agreement with our hypothesis, where we mentioned that those unknown classes help the model learn discriminative features better for very low alphas.
\section{Experiments}

Finally, we want to evaluate the model's accuracy on the test set. We will compare the baseline FedAvg algorithm to our extension proposal, named FedOS in this context, and two alternative client update aggregation schemes from literature, FedIR, and FedProx. Our experiments are limited to the more challenging values of $\alpha$, namely $\alpha=1$ and $\alpha=0.01$. All other hyperparameters remain unchanged. We split the dataset among $20$ clients according to the distribution defined by $\alpha$ and use this dataset for all $4$ algorithms. Training is done over $125$, $250$, $375$, and $500$ rounds respectively, picking $20\%$ of the clients each round to train for $1$ local epoch. The model's weights used for evaluation are those that achieved the highest accuracy on a validation dataset during the training process. We choose this approach due to the aforementioned instability while training for low values of $\alpha$. Our extension is implemented with DCGAN \cite{https://doi.org/10.48550/arxiv.1511.06434}, down-scaling the GAN's width so that it is close in size to our inference model, LeNet-5. The GAN network is trained for 90 epochs on 50\% of randomly selected Tiny Imagenet samples. For our extension we set $\mathcal{W_{U}}=1.5$, $\mathcal{F_{U}}=0.4$ for $\alpha=1$ and $\mathcal{W_{U}}=1$, $\mathcal{F_{U}}=0.8$ for $\alpha=0.01$. For FedProx we use the standard parameter setting of $\mu=1$. The results of these experiments are reported in \cref{tab:results}.

FedOS Outperforms all methods for $\alpha=1$. Sometimes even using much fewer training rounds, as in the case of $375$ rounds for FedOS and $500$ rounds for all other methods. In the case of $\alpha=0.01$, the FedOS takes more rounds to reach good results, however, it does surpass FedIR at 375 and 500 rounds. Note that since for each round we are training on only 20\% out of a total of 20 clients, only 4 clients are chosen to report their weight updates. In the case where each client only has a single class, the entire model trains on 4 each round. Thus, an interesting alternative study would be to increase the total number of clients, with each client having fewer samples. \label{sec:experiments}
   
\section{Conclusion}
In this work, we explored Federated Learning by having a look at some of the challenges that come with it, specifically training on non-IID client datasets. Furthermore, we discussed several techniques to tackle these challenges and compared their relative performance. Finally, we present our own idea, to add unknown class samples during the training process to help the local model focus on learning to distinguish features from those classes, that are present on the client device. We show that our approach performs well for highly heterogeneous clients, as well as slightly boosts performance in more homogeneous test cases. Since both the quantity and quality of our experiments were limited by the computational power provided by Colab, we would like to both further validate our hypotheses in the future, as well as compare our novel approach to more state of the art methods.

{\small
\bibliographystyle{ieee_fullname}
\bibliography{PaperForReview}
}

\end{document}